\documentclass[11pt,a4paper]{article}
\usepackage[hyperref]{naaclhlt2019}
\usepackage{times}
\usepackage{latexsym}

\usepackage{url}

\usepackage{graphicx}
\usepackage{color, soul, multirow}
\usepackage{subfig}
\usepackage{bm}
\usepackage{verbatim}
\usepackage{amssymb}
\usepackage{pifont}
\usepackage{float}
\usepackage{amsmath}

\aclfinalcopy 
\def\aclpaperid{1783} 

\title{Acoustic-to-Word Models with Conversational Context Information}

\author{Suyoun Kim$^1$ and Florian Metze$^2$ \\
  $^1$Electrical \& Computer Engineering\\
  $^2$Language Technologies Institute, School of Computer Science\\
  Carnegie Mellon University \\
  {\tt \{suyoung1, fmetze\}@andrew.cmu.edu} 
  }

\date{}

\begin{document}
\maketitle
\begin{abstract}
Conversational context information, higher-level knowledge that spans across sentences, can help to recognize a long conversation. However, existing speech recognition models are typically built at a sentence level, and thus it may not capture important conversational context information. The recent progress in end-to-end speech recognition enables integrating context with other available information (e.g., acoustic, linguistic resources) and directly recognizing words from speech. In this work, we present a direct acoustic-to-word, end-to-end speech recognition model capable of utilizing the conversational context to better process long conversations. We evaluate our proposed approach on the Switchboard conversational speech corpus and show that our system outperforms a standard end-to-end speech recognition system. 
\end{abstract}

\section{Introduction}
Many real-world speech recognition applications, including teleconferencing, and AI assistants, require recognizing and understand long conversations. In a long conversation, there exists the tendency of semantically related words or phrases re-occur across sentences, or there exists topical coherence. Thus, such conversational context information, higher-level knowledge that spans across sentences, provides important information that can improve speech recognition. However, the long conversations typically split into short sentence-level audios to make building speech recognition models computationally feasible in current state-of-the-art recognition systems \citep{xiong2017microsoft, saon2017english}.

Over the years, there have been many studies have attempted to inject a longer context information into language models. Based on a recurrent neural network (RNNs) language models \citep{mikolov2010recurrent}, \citep{mikolov2012context, wang2015larger, ji2015document, liu2017dialog, xiong2018session}, proposed using a context vector that would encode the longer context information as an additional network input. However, all of these models have been developed on text data, and therefore, it must still be integrated with a conventional acoustic model which is built separately without a longer context information, for speech recognition on long conversations.

Recently, new approaches to speech recognition models integrate all available information (e.g. acoustic, linguistic resources) in a so-called end-to-end manner proposed in \citep{graves2006connectionist, graves2014towards, hannun2014deep, miao2015eesen, bahdanau2014neural, chorowski2014end, chorowski2015attention, chan2015listen, kim2017joint}. In these approaches, a single neural network is trained to recognize graphemes or even words from speech directly. Especially, the model using semantically meaningful units, such as words or sub-word \citep{sennrich2015neural}, rather than graphemes have been showing promising results \citep{audhkhasi2017direct, li2018advancing, soltau2016neural, zenkel2017subword, palaskar2018acoustic, sanabria2018hierarchical, rao2017exploring, zeyer2018improved}.

In this work, motivated by such property of the end-to-end speech recognition approaches, we propose to integrate conversational context information within a direct acoustic-to-word, end-to-end speech recognition to better process long conversations. Thus far, the research in speech recognition systems has focused on recognizing sentences and to the best of our knowledge, there have been no studies of word-based models incorporating conversational context information. There has been recent work attempted to use the conversational context information from the preceding graphemes \citep{kim2018dialog}, however, it is limited to encode semantically meaningful context representation. Another recent work attempted to use a context information  \citep{pundak2018deep}, however, their method requires a list of phrases at inference (i.e. personalized contact list). We evaluate our proposed approach on the Switchboard conversational speech corpus \citep{swbd, godfrey1992switchboard}, and show that our model outperforms the sentence-level end-to-end speech recognition model.

\section{Models}
\label{sec:review}
\subsection{Acoustic-to-Words Models}
We perform end-to-end speech recognition using a joint CTC/Attention-based approach \citep{kim2017joint, watanabe2017hybrid}. The neural network is trained by both CTC \citep{graves2006connectionist} and Attention-based sequence-to-sequence (seq2seq) objectives \citep{bahdanau2014neural} to combine the strength of the two. With CTC, it preserves left-right order between input and output and with attention-based seq2seq, it learns the language model jointly without relying on the conditional independence assumption.

As an output, we use word-level symbols which generated from the bite-pair encoding (BPE) algorithm \citep{sennrich2015neural}. This method creates the target units based on the frequency of occurrence in training sets. Similar to \citep{zeyer2018improved, palaskar2018acoustic, sanabria2018hierarchical}, we use BPE-10k which contains roughly 10k units (9,838), including 7,119 words and 2719 sub-words.

\begin{figure}[!h]
\begin{minipage}[b]{1.0\linewidth}
  \centering
  \centerline{\includegraphics[width=8.5cm]{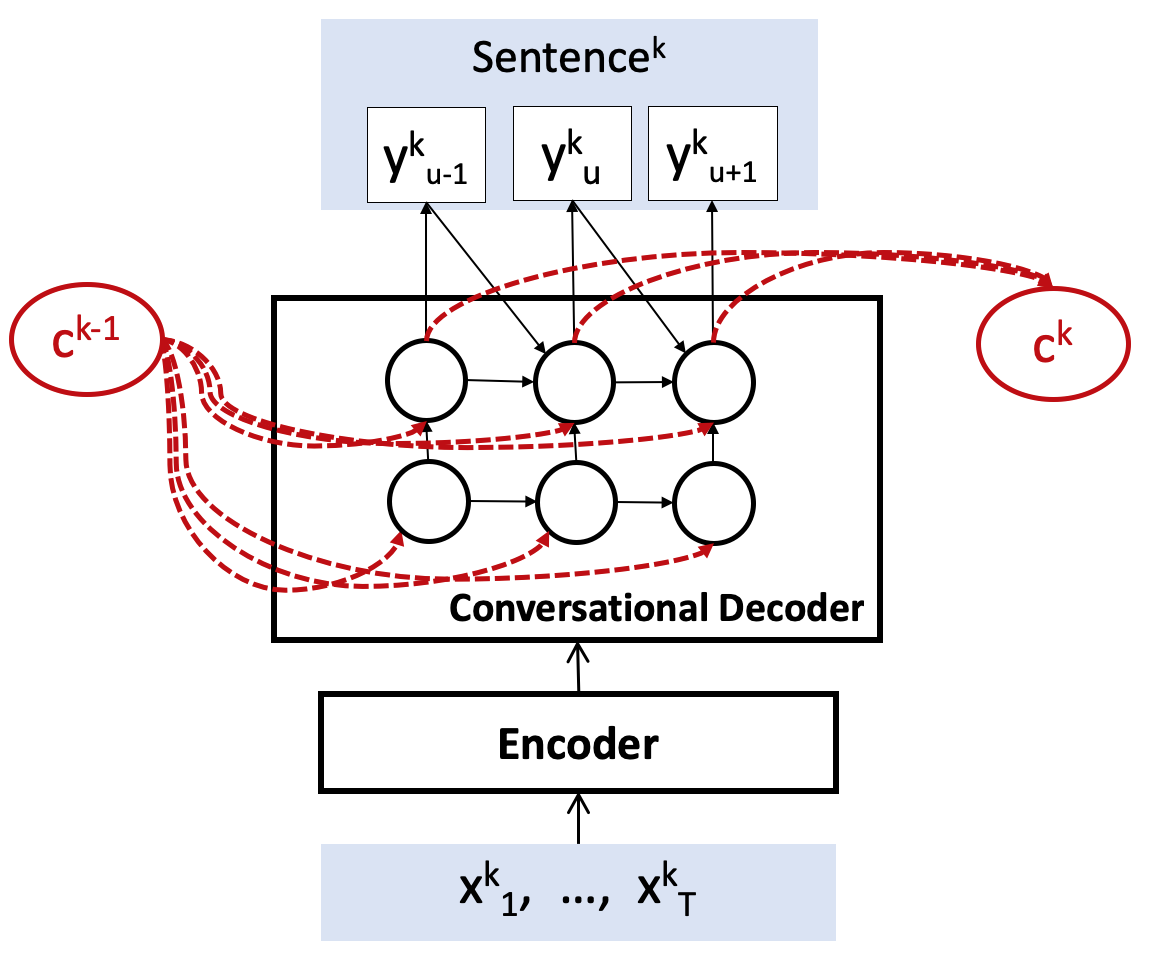}}
\end{minipage}
\caption{The architecture of our end-to-end speech recognition model with conversational context information. The $c^{k-1}$ is the conversational context vector generated from the preceding $k-1$ sentence red curved dashed line represents the context information flow within the same conversation.}
\label{fig:gate_function_analysis}
\end{figure}

\subsection{Conversational Context Representation}
\label{sec:dcrnnlm}
In order to use conversational context information within the end-to-end speech recognition framework, we extend the decoder sub-network to predict the output additionally conditioning on conversational context. To do so, we encode the preceding sentence into a single vector, a conversational context vector, then inject to decoder network as an additional input at every output step. 

Let we have K sentences in a conversation. For $k$-th sentence, $s^k$, we have $T^k$-length input acoustic feature ($\bm{x^k}$) and $U^k$-length output words. Our proposed decoder generates the probability distribution over words ($ ~ y_u^k$),   conditioned on 1) high-level representation ($\bm{h^k}$) of input ($\bm{x^k}$) generated from encoder, and 2) all the words seen previously ($y^k_{1:u-1}$), and 3) previous decoder state ($d^k_{u-1}$)  4) additionally conditioning on conversational context vector ($c^{k-1}$), which represents the information of the preceding sentence ($k-1$):
    \begin{align}
		\bm{h^k}       = & \textit{Encoder}(\bm{x^k}) \\
	    y^k_u \sim & \textit{Decoder}(\bm{h^k}, y^k_{1:u-1}, d^k_{u-1}, c^{k-1})
	\end{align}	

We represent the context vector, $c^{k-1}$, from the preceding sentence in two different ways: (a) mean of word embedding, and (b) attentional word embedding. We first generate one-hot word vectors, and then we simply take the mean over word vectors to obtain a single vector in method (a), or we use attention mechanism over word vectors to obtain the weight over the words and then perform the weighted-sum. The parameter of the attention mechanism is optimized towards minimizing the conversation ID classification error similar to \citep{kim2018dialog}. The context vector is merged with a decoder state at every output step as follows:
\begin{align}
     \Hat{d}^k_{u-1} &= \tanh (W d^k_{u-1} + V c^{k-1} + b) \\
     y^k_u  & \sim \text{softmax}(\text{LSTM}( \Hat{d}^k_{u-1} , \bm{h^k_u}, y^k_{1:u-1})))
\end{align}
where $W, V, b$ are trainable parameters. 

In order to learn and use the conversational-context during training and decoding, we serialize the
sentences based on their onset times and their conversations rather than the random shuffling of data.
We shuffle data at the conversation level and create mini-batches that contain only one sentence of each conversation.

\section{Experiments}
\label{sec:exp}
\subsection{Datasets}

We investigated the performance of the proposed model on the Switchboard LDC corpus (97S62) which has a 300 hours training set. We split the Switchboard data into two groups, then used 285 hours of data (192 thousand sentences) for model training and 5 hours of data (4 thousand sentences) for hyper-parameter tuning. The evaluation was carried out on the HUB5 Eval 2000 LDC corpora (LDC2002S09, LDC2002T43), which have 3.8 hours of data (4.4 thousand sentences), and we show separate results for the Callhome English (CH) and Switchboard (SWB) evaluation sets. We denote train\_nodup, train\_dev, SWB, and CH as our training, development, and two evaluation datasets for CH and SWB, respectively. There are 2,402 conversations in training sets and 20 conversations in CH, and 20 conversations in SWB. 

We sampled all audio data at 16kHz, and extracted 80-dimensional log-mel filterbank coefficients with 3-dimensional pitch features, from 25~ms frames with a 10ms frame shift. We used 83-dimensional feature vectors to input to the network in total. We used 9,840 distinct labels: 9,838 word-level BPE units, start-of-speech/end-of-speech, and blank tokens. Note that no pronunciation lexicon was used in any of the experiments.

\begin{table*}[t]
\centering
\caption{Comparison of word error rates (WER) on Switchboard 300h with standard end-to-end speech recognition models and our proposed end-to-end speech recognition models with conversational context.}
\label{tab:result}
\resizebox{\textwidth}{!}{
\begin{tabular}{|r|r|r|r|r|}
\hline
Model   & Output Units   & LM & SWB (WER \%) & CH (WER \%) \\
\hline
\hline
\textbf{Prior Models} & & & & \\
LF-MMI \citep{povey2016purely} & context-dependend phones & O & 9.6 & 19.3 \\
CTC \citep{zweig2017advances} & Char & O & 19.8 & 32.1 \\
CTC \citep{audhkhasi2017building} & Word (Phone init.) & O & 14.6  & 23.6 \\
CTC \citep{sanabria2018hierarchical} & Char, BPE-\{300, 1k, 10k\} & O & 12.5 &23.7 \\
Seq2Seq \citep{palaskar2018acoustic} & BPE-10k & O & 21.3 & 35.7 \\
Seq2Seq \citep{zeyer2018improved} & BPE-1k & O & 11.8 & 25.7 \\
\hline
\hline
\textbf{Our Sentence-level Baseline} & & & & \\
Our baseline & BPE-10k & x & 17.6 & 30.6 \\
Our baseline & BPE-10k & O (only swb) & 17.0 & 29.7 \\
\hline
\textbf{Our Proposed Conversational Model} & & & & \\
w/ Context \textit{(a) mean} & BPE-10k & O (only swb) & 16.3 & 29.0 \\
w/ Context \textit{(b) att} & BPE-10k & O (only swb) & 16.4 & 29.2 \\
w/ Context \textit{(b) att} + pre-training & BPE-10k & O (only swb) & 16.0 & 28.9  \\
\hline
\end{tabular}
}
\end{table*}




\begin{table}[t]
\caption{ Perplexities on a held-out set of our proposed conversational context LM and baselines.}
\label{tab:perp}
\begin{center}
\resizebox{\columnwidth}{!}{
\begin{tabular}{|r|r|r|r|}
\hline
Models & Fisher text & PPL        \\
\hline
\hline
Baseline LM     & x      & 74.15 \\
Baseline LM     & o & 72.81 \\
\hline
Proposed Conversational LM      & x      & 67.03 \\
Proposed Conversational LM      & o & 64.30 \\
\hline
\end{tabular}
}
\end{center}
\end{table}

\subsection{Training and decoding}
We used joint CTC/Attention end-to-end speech recognition architecture \citep{kim2017joint, watanabe2017hybrid} with ESPnet toolkit \citep{watanabe2018espnet}. We used a CNN-BLSTM encoder as suggested in \citep{zhang2017very, hori2017advances}. We followed the same six-layer CNN architecture as the prior study, except we used one input channel instead of three since we did not use delta or delta delta features. Input speech features were downsampled to (1/4 x 1/4) along with the time-frequency axis. Then, the 6-layer BLSTM with 320 cells was followed by CNN. We used a location-based attention mechanism \citep{chorowski2015attention}, where 10 centered convolution filters of width 100 were used to extract the convolutional features. 

The decoder network of both our proposed models and the baseline models was a 2-layer LSTM with 300 cells. Our proposed models additionally require linear projection layer in order to encode the conversational context vector and merge with decoder states. 

We also built an external RNN-based language model (RNNLM) on the same BPE-10k sets on the same Switchboard transcriptions. The RNNLM network architecture was a two-layer LSTM with 650 cells. This network was used only for decoding. 

The AdaDelta algorithm \citep{zeiler2012adadelta} with gradient clipping \citep{pascanu2013difficulty} was used for optimization. We used $\lambda = 0.5$ for joint CTC/Attention training. We bootstrap the training our proposed conversational end-to-end models from the baseline end-to-end models. When we decode with RNNLM, we used joint decoder which combines the output label scores from the AttentionDecoder, CTC, and RNNLM by using shallow fusion \citep{hori2017advances}: 
\begin{align}
\begin{split}
    \bm{y}* = \text{argmax} \{ & \log p_{att}(\bm{y}|\bm{x}) \\
                      &+ \alpha \log p_{att}(\bm{y}|\bm{x}) \\
                      &+ \beta \log p_{rnnlm}(\bm{y}) \}
\end{split}
\end{align}
The scaling factor of CTC, and RNNLM scores were $\alpha = 0.3$, and $\beta = 0.3$, respectively. We used a beam search algorithm similar to \citep{sutskever2014sequence} with the beam size 10 to reduce the computation cost. We adjusted the score by adding a length penalty, since the model has a small bias for shorter utterances. The final score $s(\bm{y}|\bm{x})$ is normalized with a length penalty $0.5$.

The models were implemented by using the PyTorch deep learning library \citep{paszke2017automatic}, and ESPnet toolkit \citep{kim2017joint, watanabe2017hybrid, watanabe2018espnet}.

\section{Results}
\label{sec:result}

We evaluated both the end-to-end speech recognition model which was built on sentence-level data and our proposed end-to-end speech recognition model which leveraged conversational context information.

Table \ref{tab:result} shows the WER of our baseline, proposed models, and several other published results those were only trained on 300 hours Switchboard training data. As shown in Table \ref{tab:result}, we obtained a performance gain over our baseline by using the conversational context information. Our proposed model \textit{(a) mean} shows 4.1\% and 2.4\% relative improvement over our baseline on SWB and CH evaluation set, respectively. Our proposed model \textit{(b) att} shows 3.5\% and 1.7\% relative improvement over our baseline on SWB and CH evaluation set, respectively. We also found that we can obtain further accuracy improvement by pre-training the decoder part only with transcription. With this pre-training technique, the \textit{(b) att} shows 5.9\% and 2.7\% relative improvement. Unlike the previous work \citep{renduchintala2018multi}, we did not use any additional encoder for the text data.  

We also build the language model with or without the conversational context information. Table \ref{tab:perp} shows the perplexity on a held-out set of our baseline LM and our conversational LM. We observed that incorporating the conversational context improves performance showing that 9.6\% and 11.7\% relative improvement on \textit{SWBD only} and \textit{SWBD + Fisher}. Note that the Fisher (LDC2004T19) parts \citep{cieri2004fisher} of transcriptions is only used in these experiments. 
 
\begin{figure}[!h]
\begin{minipage}[b]{1.0\linewidth}
  \centering
  \centerline{\includegraphics[width=8.5cm]{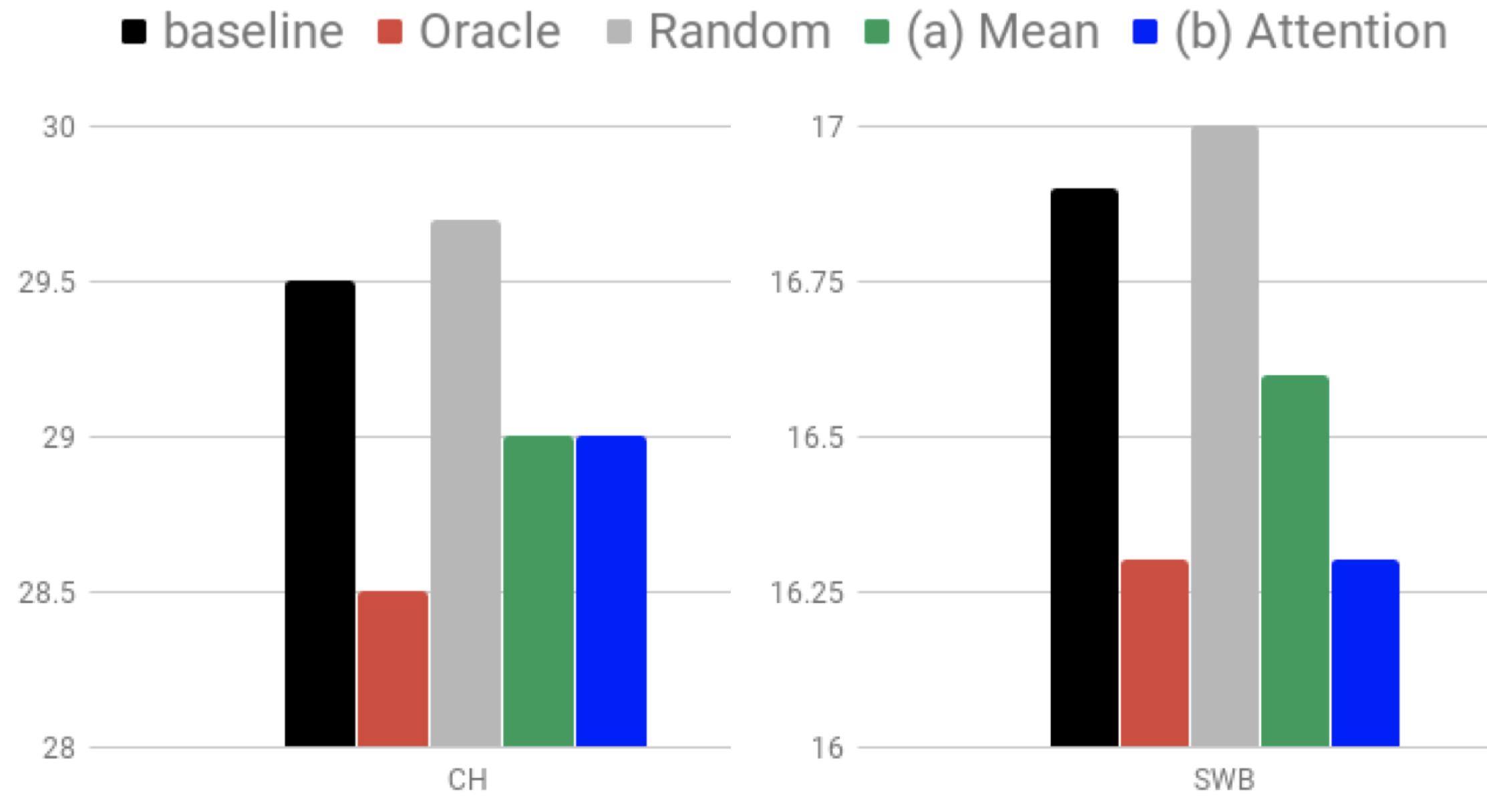}}
\end{minipage}
\caption{The architecture of our end-to-end speech recognition model with conversational context information. The $c^{k-1}$ is the conversational context vector generated from the preceding $k-1$ sentence red curved line represents the context information flow within the same conversation.}
\label{fig:oracle_random}
\end{figure}

We performed analyses in order to verify the conversational vector helps to improve recognition accuracy. We generate the context vector from an oracle preceding sentence and a random sentence, in addition to our predicted sentence. As described in Figure \ref{fig:oracle_random}, the model using the oracle context performed best and the model using the random context was even worse than the baseline. Our model outperformed over the baseline and the model using the random context, we can conclude that the benefit from our proposed method is coming from the conversational context information.

\section{Conclusion}
\label{sec:conclusion}
We proposed an acoustic-to-word model capable of utilizing the conversational context to better process long conversations. A key aspect of our model is that the whole system can be trained with conversational context information in an end-to-end framework. Our model was shown to outperform previous end-to-end speech recognition models trained on isolated utterances by incorporating preceding conversational context representations. 

\section*{Acknowledgments}
We gratefully acknowledge the support of NVIDIA Corporation with the donation of the Titan Xp GPU used for this research. This work also used the Bridges system, which is supported by NSF award number ACI-1445606, at the Pittsburgh Supercomputing Center (PSC). This research was supported by a fellowship from the Center for Machine Learning and Health (CMLH) at Carnegie Mellon University. 

\bibliography{naaclhlt2019}
\bibliographystyle{acl_natbib}

\end{document}